\documentclass[10pt,twocolumn,letterpaper]{article}

\usepackage{cvpr}
\usepackage{times}
\usepackage{epsfig}
\usepackage{graphicx}
\usepackage{amsmath}
\usepackage{amssymb}
\usepackage{mathtools}

\newcommand\blfootnote[1]{%
  \begingroup
  \renewcommand\thefootnote{}\footnote{#1}%
  \addtocounter{footnote}{-1}%
  \endgroup
}

\usepackage[pagebackref=true,breaklinks=true,letterpaper=true,colorlinks,bookmarks=false]{hyperref}

\cvprfinalcopy 

\def\cvprPaperID{2857}

\ifcvprfinal\fi

\title{Estimating 3D Motion and Forces of Person-Object Interactions\\ from Monocular Video}

\author{Zongmian Li\textsuperscript{1,2} \qquad\qquad Jiri Sedlar\textsuperscript{3} \qquad\qquad Justin Carpentier\textsuperscript{1,2} \\ Ivan Laptev\textsuperscript{1,2} \qquad\qquad Nicolas Mansard\textsuperscript{4} \qquad\qquad Josef Sivic\textsuperscript{1,2,3}\\
\textsuperscript{1}DI ENS PSL \qquad\qquad \textsuperscript{2}Inria \qquad\qquad \textsuperscript{3}CIIRC, CTU in Prague \qquad\qquad \textsuperscript{4}LAAS-CNRS
}

\begin{document}
\sloppy 

\maketitle

\blfootnote{\textsuperscript{*}Please see our project webpage~\cite{project-page} for trained models, data and code. }
\blfootnote{\textsuperscript{1}D\'{e}partement d'informatique de l'ENS, \'{E}cole normale sup\'{e}rieure, CNRS, PSL Research University, 75005 Paris, France.}
\blfootnote{\textsuperscript{3}Czech Institute of Informatics, Robotics and Cybernetics at the Czech Technical University in Prague, Czech Republic.}

\begin{abstract}
    In this paper, we introduce a method to automatically reconstruct the 3D motion of a person interacting with an object from a single RGB video.
    Our method estimates the 3D poses of the person and the object, contact positions, and forces and torques actuated by the human limbs. 
    The main contributions of this work are three-fold. 
    First, we introduce an approach to jointly estimate the motion and the actuation forces of the person on the manipulated object by modeling contacts and the dynamics of their interactions. 
    This is cast as a large-scale trajectory optimization problem.
    Second, we develop a method to automatically recognize from the input video the position and timing of contacts between the person and the object or the ground, thereby significantly simplifying the complexity of the optimization. 
    Third, we validate our approach on a recent MoCap dataset with ground truth contact forces and demonstrate its performance on a new dataset of Internet videos showing people manipulating a variety of tools in unconstrained environments.
\end{abstract}


\begin{figure}[t]
\includegraphics[width=\linewidth]{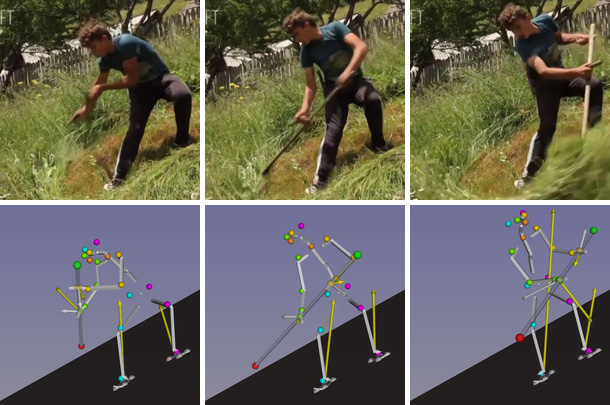}
\vspace{-4mm}
\caption{\small Our method automatically reconstructs in 3D the object manipulation action captured by a single RGB video.
    \textbf{Top:} Frames from the input video. 
    \textbf{Bottom:} the output human and object 3D motion including the recovered contact forces (yellow arrows) and moments (white arrows).} 
\label{fig:teaser}
\vspace{-3mm}
\end{figure}

\section{Introduction}\label{sec:introduction}
People can easily learn how to break concrete with a sledgehammer or cut hay using a scythe by observing other people performing such tasks in instructional videos, for example.
They can also easily perform the same task in a different context.
This involves advanced visual intelligence capabilities such as recognizing and interpreting complex person-object interactions that achieve a specific goal. 
Understanding such complex interactions is a key to building autonomous machines that learn how to interact with the physical world by observing people.

This work makes a step in this direction and describes a method to estimate the 3D motion and actuation forces of a person manipulating an object given a single unconstrained video as input, as shown in Figure~\ref{fig:teaser}.
This is an extremely challenging task.
First, there are inherent ambiguities in the 2D-to-3D mapping from a single view: multiple 3D human poses correspond to the same 2D input.
Second, human-object interactions often involve contacts, resulting in discontinuities in the motion of the object and the human body part in contact. For example, one must place a hand on the hammer handle before picking the hammer up. The contact motion strongly depends on the physical quantities such as the mass of the object and the contact forces exerted by the hand, which renders modeling of contacts a very difficult task.
Finally, the tools we consider in this work, such as hammer, scythe, or spade, are particularly difficult to recognize due to their thin structure, lack of texture, and frequent occlusions by hands and other human parts. 

To address these challenges, we propose a method to jointly estimate the 3D trajectory of both the person and the object by visually recognizing contacts in the video and modeling the dynamics of the interactions.
We focus on rigid stick-like hand tools (e.g.~hammer, barbell, spade, scythe) with no articulation and approximate them as 3D line segments.
Our key idea is that, when a human joint is in contact with an object, the object can be integrated as a constraint on the movement of the human limb.
For example, the hammer in Figure~\ref{fig:teaser} provides a constraint on the relative depth between the person's two hands.
Conversely, 3D positions of the hands in contact with the hammer provide a constraint on the hammer's depth and 3D rotation.
To deal with contact forces, we integrate physics in the estimation by modeling dynamics of the person and the object.
Inspired by recent progress in humanoid locomotion research~\cite{carpentier2018multi}, we formulate person-object trajectory estimation as an optimal control problem given the contact state of each human joint.
We show that contact states can be automatically recognized from the input video using a deep neural network.

\section{Related work}
Here we review the key areas of related work in both computer vision and robotics literature.

\noindent \textbf{Single-view 3D pose estimation } aims to recover the 3D joint configuration of the person from the input image.
Recent human 3D pose estimators either attempt to build a \textit{direct mapping} from image pixels to the 3D joints of the human body or break down the task into \textit{two stages}: estimating pixel coordinates of the joints in the input image and then lifting the 2D skeleton to 3D.
Existing direct approaches either rely on generative models to search the state space for a plausible 3D skeleton that aligns with the image evidence \cite{sidenbladh2000stochastic,gammeter2008articulated,gall2010optimization} or, more recently, extract deep features from images and learn a discriminative regressor from the 2D image to the 3D pose \cite{hmrKanazawa18,moreno20173d,pavlakos2017coarse,tekin2016direct}.
Building on the recent progress in 2D human pose estimation \cite{newell2016stacked,newell2017associative,insafutdinov2016deepercut,cao2017realtime}, two-stage methods have been shown to be very effective \cite{akhter2015pose,zhou2016sparseness,bogo2016keep,chen20173d} and achieve state-of-the-art results \cite{martinez2017simple} on 3D human pose benchmarks~\cite{h36m_pami}.
To deal with depth ambiguities, these estimators rely on good pose priors, which are either hand-crafted or learnt from large-scale MoCap data~\cite{zhou2016sparseness,bogo2016keep,hmrKanazawa18}. However, unlike our work, these methods do not consider explicit models for 3D person-object interactions with contacts.

\noindent \textbf{Understanding human-object interactions } involves both recognition of actions and  modeling of interactions.
In action recognition, most existing approaches that model human-object interactions do not consider 3D, instead model interactions and contacts in the 2D image space~\cite{gupta2009observing,delaitre2011learning,yao2012recognizing,prest2013explicit}.
Recent works in scene understanding~\cite{jiang2013hallucinated,fouhey2014people} consider interactions in 3D but have focused on static scene elements rather than manipulated objects as we do in this work.
Tracking 3D poses of people interacting with the environment has been demonstrated for bipedal walking~\cite{brubaker2007physics,brubaker2009estimating} or in sports scenarios~\cite{videomocap2010}.
However, these works do not consider interactions with objects. Furthermore,~\cite{videomocap2010} requires manual annotation of the input video.

There is also related work on modeling person-object interactions in robotics~\cite{tassa2012} and computer animation~\cite{boulic1990}.
Similarly to people, humanoid robots interact with the environment by creating and breaking contacts~\cite{herdt2010}, for example, during walking.
Typically, generating artificial motion is formulated as an optimal control problem, transcribed into a high-dimensional numerical optimization problem, seeking to minimize an objective function under contact and feasibility constraints~\cite{diehl2006,schultz2010modeling}.
A known difficulty is handling the non-smoothness of the resulting optimization problem introduced by the creation and breaking of contacts~\cite{westervelt2003hybrid}.
Due to this difficulty, the sequence of contacts is often computed separately and not treated as a decision variable in the optimizer~\cite{kuffner2005motion,tonneau2018}.
Recent work has shown that it may be possible to decide both the continuous movement and the contact sequence together, either by implicitly formulating the contact constraints~\cite{posa2014direct} or by using invariances to smooth the resulting optimization problem~\cite{mordatch2012discovery,winkler2018gait}.

In this paper, we take advantage of rigid-body models introduced in robotics and formulate the problem of estimating 3D person-object interactions from monocular video as an optimal control problem under contact constraints.
We overcome the difficulty of contact irregularity by first identifying the contact states from the visual input, and then localizing the contact points in 3D via our trajectory estimator.
This allows us to treat multi-contact sequences (like walking) without manually annotating the contact phases.

\noindent \textbf{Object 3D pose estimation} methods often require depth or RGB-D data as input~\cite{tejani2014latent,doumanoglou2016,hinterstoisser2016going}, which is restrictive since depth information is not always available (e.g.~for outdoor scenes or specular objects), as is the case of our instructional videos. 
Recent work has also attempted to recover object pose from RGB input only~\cite{brachmann2016uncertainty,rad2017bb8,posecnn2017,deepim2018eccv,oberweger2018eccv,grabner2018cvpr,rad2018cvpr}.
However, we found that the performance of these methods is limited for the stick-like objects we consider in this work.
Instead, we recover the 3D pose of the object via localizing and segmenting the object in 2D, and then jointly recovering the 3D trajectory of both the human limbs and the object. As a result, both the object and the human pose help each other to improve their joint 3D trajectory by leveraging the contact constraints.

\noindent \textbf{Instructional videos.}
Our work is also related to recent efforts in learning form Internet instructional videos \cite{malmaud2015s,Alayrac16unsupervised, Alayrac16unsupervised} that aim to segment input videos into clips containing consistent actions. 
In contrast, we focus on extracting a detailed representation of the object manipulation in the form of a 3D person-object trajectory with contacts and underlying manipulation forces.

\begin{figure*}[t]
    \centering
    \includegraphics[width=\textwidth]{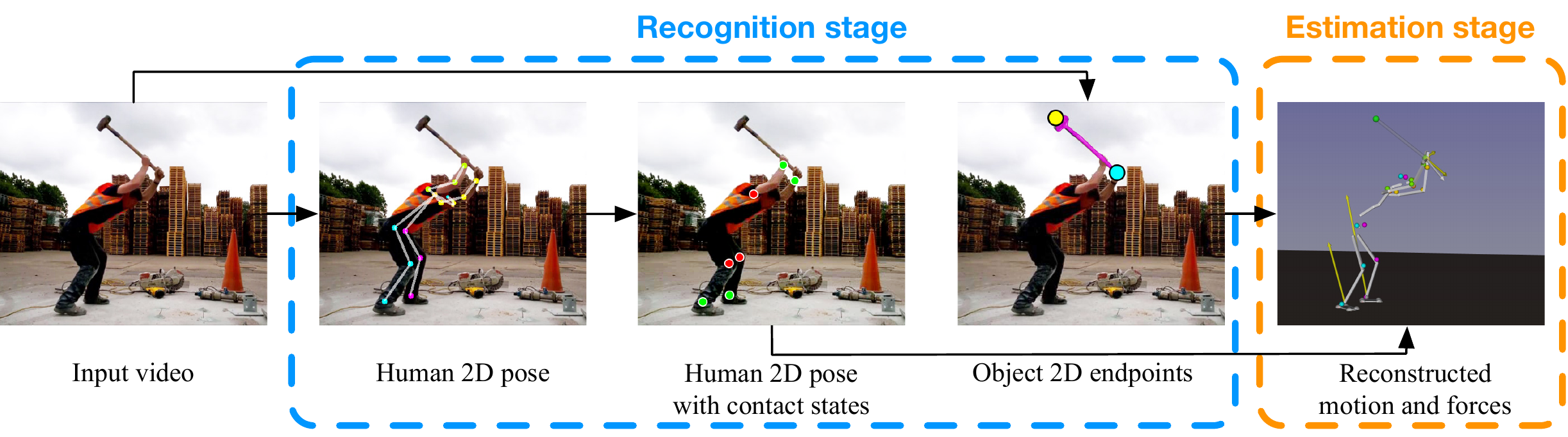}
    \caption{\small Overview of the proposed method. In the recognition stage, the system estimates from the input video the person's 2D joints, the hammer's 2D endpoints and the contact states of the individual joints.
    The human joints and the object endpoints are visualized as colored dots in the image.
    Human joints recognized as in contact are shown in green, joints not in contact in red.
    In the estimation stage, these image measurements are fused in a trajectory estimator to recover the human and object 3D motion together with the contact positions and forces.
    }
    \vspace*{-3mm}
    \label{fig:system_model_scheme_1}
\end{figure*}

\section{Approach overview}
We are given a video clip of a person manipulating an object or in another way interacting with the scene. 
Our approach, illustrated in Figure~\ref{fig:system_model_scheme_1}, receives as input a sequence of frames and automatically outputs the 3D trajectories of the human body, the manipulated object, and the ground plane. At the same time, it localizes the contact points and recovers the contact forces that actuate the motion of the person and the object.
Our approach proceeds along two stages. 
In the first, \textit{recognition stage}, we extract 2D measurements from the input video. These consist of 2D locations of human joints, 2D locations of a small number of predefined object endpoints, and contact states of selected joints over the course of the video. 
In the second, \textit{estimation stage}, these image measurements are then fused in order to estimate the 3D motion, 3D contacts, and the controlling forces of both the person and the object.
The person and object trajectories, contact positions, and contact forces are constrained jointly by our carefully designed contact motion model, force model, and dynamics equations.
Finally, the reconstructed object manipulation sequence can be applied to control a humanoid robot via behavior cloning.

In the following, we start in Section~\ref{sec:main_stage} by describing the estimation stage giving details of the formulation as an optimal control problem.
Then, in Section \ref{sec:extract_2d_measurements} we give details of the recognition stage including 2D human pose estimation, contact recognition, and object 2D endpoint estimation.
Finally, we describe results in Section \ref{sec:results}.

\section{Estimating person-object trajectory under contact and dynamics constraints}\label{sec:main_stage} 
We assume that we are provided with a video clip of duration $T$ depicting a human subject manipulating an object. We encode the 3D poses of the human and the object, including joint translations and rotations, in the configuration vectors $q^\mathrm{h}$ and $q^\mathrm{o}$, for the human and the object respectively.
We define a constant set of $K$ contact points between the human body and the object (or the ground plane).
Each contact point corresponds to a human joint, and is activated whenever that human joint is recognized as in contact.
At each contact point, we define a contact force $f_k$, whose value is non-zero whenever the contact point $k$ is active.
The state of the complete dynamical system is then obtained by concatenating the human and the object joint configurations $q$ and velocities $\dot{q}$ as 
$x \coloneqq \left(q^\mathrm{h}, q^\mathrm{o}, \dot{q}^\mathrm{h}, \dot{q}^\mathrm{o}\right)$.
Let $\tau^\mathrm{h}_\mathrm{m}$ be the joint torque vector describing the actuation  by human muscles. This is a $n_q-6$ dimensional vector where $n_q$ is the dimension of the human body configuration vector. We define the control variable $u$ as the combination of the joint torque vector together with contact forces at the $K$ contact joints, $u \coloneqq \left(\tau^\mathrm{h}_\mathrm{m}, f_k, k=1,...,K\right)$.
To deal with sliding contacts, we further define a contact state $c$ that consists of the relative positions of all the contact points with respect to the object (or ground) in the 3D space.

Our goal is two-fold. We wish to (i) estimate smooth and consistent human-object and contact trajectories $\underline{x}$ and $\underline{c}$, while (ii) recovering the control $\underline{u}$ which gives rise to the observed motion\footnote{In this paper, trajectories are denoted as underlined variables, e.g. $\underline{x},\underline{u}~\text{or}~\underline{c}$.}.
This is achieved by jointly optimizing the 3D trajectory $\underline{x}$, contacts $\underline{c}$, and control $\underline{u}$ given the measurements (2D positions of human joints and object endpoints together with contact states of human joints) obtained from the input video. 
The intuition is that the human and the object's 3D poses should match their respective projections in the image while their 3D motion is linked together by the recognized contact points and the corresponding contact forces.
In detail, we formulate person-object interaction estimation as an optimal control problem with contact and dynamics constraints:
\begin{align}
    \underset{\underline{x},\underline{u},\underline{c}}{\text{minimize}} &\quad \sum_{e \in \{\mathrm{h}, \mathrm{o}\}}{\int_{0}^{T}{l^{e}\left(x, u, c\right)\mathrm{d}t}}, \label{eq:general_problem}\\
    \text{subject to} &\quad  \kappa(x, c) = 0 \quad \text{(contact motion model)}, \label{eq:contact_motion_model}\\
    & \quad \dot{x} = f\left(x, c, u\right)  \quad \text{(full-body dynamics)}, \label{eq:full_body_dynamics}\\
    &\quad u \in \mathcal{U}\quad \text{(force model)},\label{eq:force_model}
\end{align}
where $e$ denotes either `$\mathrm{h}$' (human) or `$\mathrm{o}$' (object), and the constraints \eqref{eq:contact_motion_model}-\eqref{eq:force_model} must hold for all $t\in[0,T]$.
The loss function $l^e$ is a weighted sum of multiple costs capturing (i) the data term measuring the discrepancy between the observed and re-projected 2D joint and object endpoint positions, (ii) the prior on the estimated 3D poses, (iii) the physical plausibility of the motion and (iv) the smoothness of the trajectory.
Next, we in turn describe these cost terms as well as the insights leading to their design choices.
For simplicity, we ignore the superscript $e$ when introducing a cost term that exists for both the human $l^\mathrm{h}$ and the object $l^\mathrm{o}$ component of the loss.
We describe the individual terms using continuous time notation as used in the overall problem formulation \eqref{eq:general_problem}. 
A discrete version of the problem as well as the optimization and implementation details are relegated to Section \ref{sec:optimization}.

\subsection{Data term: 2D re-projection error}\label{sec:likelihood_of_the_3d_pose}
We wish to minimize the re-projection error of the estimated 3D human joints and 3D object endpoints with respect to the 2D measurements obtained in each video frame.
In detail, let $j=1,...,N$ be human joints or object endpoints and $p^{\mathrm{2D}}_j$ their 2D position observed in the image. 
We aim to minimize the following \textit{data term}
\begin{align}
    l_\mathrm{data} =  \sum_j{\rho\left(p^{\mathrm{2D}}_{j} - P_\mathrm{cam}(p_{j}(q))\right)} \label{eq:data},
\end{align}
where $P_\mathrm{cam}$ is the camera projection matrix and $p_j$ the 3D position of joint or object endpoint $j$ induced by the  person-object configuration vector $q$.
To deal with outliers, we use the robust Huber loss, denoted by $\rho$.

\subsection{Prior on 3D human poses}
\label{sec:pose_prior}
A single 2D skeleton can be a projection of multiple 3D poses, many of which are unnatural or impossible exceeding the human joint limits.
To resolve this, we incorporate into the human loss function $l^\mathrm{h}$ a pose prior similar to~\cite{bogo2016keep}.
The pose prior is obtained by fitting the SMPL human model \cite{loper2015smpl} to the CMU MoCap data \cite{CmuMarkerDataset} using MoSh \cite{Loper:SIGASIA:2014} and fitting a Gaussian Mixture Model (GMM) to the resulting SMPL 3D poses.
We map our human configuration vector $q^\mathrm{h}$ to a SMPL pose vector $\theta$ and compute the likelihood under the pre-trained GMM
\begin{align}
l^\mathrm{h}_\mathrm{pose} &= -\log\left(p(q^\mathrm{h}; \text{GMM})\right).\label{eq:human_pose}
\end{align}
During optimization, $l^\mathrm{h}_\mathrm{pose}$ is minimized in order to favor more plausible human poses against rare or impossible ones.

\subsection{Physical plausibility of the motion}\label{sec:physical_plausibility}
Human-object interactions involve contacts coupled with interaction forces, which are not included in the data-driven cost terms \eqref{eq:data} and \eqref{eq:human_pose}.
Modeling contacts and physics is thus important to reconstruct object manipulation actions from the input video.
Next, we outline models for describing the motion of the contacts and the forces at the contact points. Finally, the contact motions and forces, together with the system state $\underline{x}$, are linked by the laws of mechanics via the dynamics equations, which constrain the estimated person-object interaction. This full body dynamics constraint is detailed at the end of this subsection.

\paragraph{Contact motions.}
In the recognition stage, our contact recognizer predicts, given a human joint (for example, left hand, denoted by $j$), a sequence of contact states $\delta_j: t \longrightarrow \{1,0\}$.
Similarly to \cite{carpentier2018multi}, we call a \textit{contact phase} any time segment in which $j$ is in contact, i.e., $\delta_j=1$.
Our key idea is that the 3D distance between human joint $j$ and the active contact point on the object (denoted by $k$) should remain zero during a contact phase:
\begin{align}
     \left\|p^\mathrm{h}_j(q^\mathrm{h}) - p^\mathrm{c}_k(x, c)\right\|=0\quad  \text{(point contact)},\label{eq:point_contact}
\end{align}
where $p^\mathrm{h}_{j}$ and $p^\mathrm{c}_k$ are the 3D positions of joint $j$ and object contact point $k$, respectively. Note that position of the object contact point $p^\mathrm{c}_k(x, c)$ depends on the state vector $x$ describing the human-object configuration and the relative position $c$ of the contact along the object. 
The position of contact $p^\mathrm{c}_k$ is subject to a feasible range denoted by $\mathcal{C}$.
For stick-like objects such as hammer, $\mathcal{C}$ is approximately the 3D line segment representing the handle. For the ground, the feasible range $\mathcal{C}$ is a 3D plane. 
In practice, we implement $p^\mathrm{c}_k\in \mathcal{C}$ by putting a constraint on the trajectory of relative contact positions $\underline{c}$.

Equation \eqref{eq:point_contact} applies to most common cases where the contact area can be modeled as a point.
Examples include the hand-handle contact and the knee-ground contact.
To model the \textit{planar contact} between the human sole and ground, we approximate each sole surface as a planar polygon with four vertices, and apply the point contact model at each vertex.
In our human model, each sole is attached to its parent ankle joint, and therefore the four vertex contact points of the sole are active when $\delta_\mathrm{ankle}=1$.

The resulting overall contact motion function $\kappa$ in problem \eqref{eq:general_problem} is obtained by unifying the point and the planar contact models:
\begin{align}
\kappa(x, c)=\sum_{j}\sum_{k\in \phi(j)}\delta_j\left\|T^{(kj)} \left(p^\mathrm{h}_j(q^\mathrm{h})\right) - p^\mathrm{c}_k(x, c)\right\|, \label{eq:contact_model}
\end{align}
where the external sum is over all human joints.
The internal sum is over the set of active object contact points mapped to their corresponding human joint $j$ by mapping $\phi(j)$.
The mapping $T^{(kj)}$ translates the position of an ankle joint $j$ to its corresponding $k$-th sole vertex; it is an identity mapping for non-ankle joints.

\paragraph{Contact forces.}
During a contact phase of the human joint $j$, the environment exerts a contact force $f_k$ on each of the active contact points in $\phi(j)$.
$f_k$ is always expressed in contact point $k$'s local coordinate frame.
We distinguish two types of contact forces: (i) 6D spatial forces exerted by objects and (ii) 3D linear forces due to ground friction.
In the case of object contact, $f_k$ is an unconstrained 6D spatial force with 3D linear force and 3D moment.
In the case of ground friction, $f_k$ is constrained to lie inside a 3D friction cone $\mathcal{K}^3$ (also known as the quadratic Lorentz ``ice-cream'' cone \cite{carpentier2018multi}) characterized by a positive friction coefficient $\mu$.
In practice, we approximate $\mathcal{K}^3$ by a 3D pyramid spanned by a basis of $N=4$ generators, which allows us to represent $f_k$ as the convex combination $f_k = \sum_{n=1}^{N}{\lambda_{kn}g^{(3)}_n}$, where $\lambda_{kn}\geq 0$ and $g^{(3)}_n$ with $n=1,2,3,4$ are the 3D generators of the contact force.
We sum the contact forces induced by the four sole-ground contact points and express a unified contact force in the ankle's frame:
\begin{align}
    f_j=\sum_{k=1}^4
    \begin{pmatrix}
        f_k \\ 
        p_k\times f_k
    \end{pmatrix}
    =\sum_{k=1}^{4}\sum_{n=1}^{N}\lambda_{jkn}g^{(6)}_{kn},
\end{align}
where $p_k$ is the position of contact point $k$ expressed in joint $j$'s (left/right ankle) frame, $\times$ is the cross product operator, $\lambda_{jkn}\geq 0$, and $g^{(6)}_{kn}$ are the 6D generators of $f_j$.
Please see Appendix \ref{appendix:ground_force_generators} for additional details including the expressions of $g^{(3)}_{n}$ and $g^{(6)}_{kn}$.

\paragraph{Full body dynamics.}
The full-body movement of the person and the manipulated object is described by the Lagrange dynamics equation:
\begin{align}
    M(q)\ddot{q} + b(q, \dot{q}) = g(q) + \tau, \label{eq:dynamics_equation}
\end{align}
where $M$ is the generalized mass matrix, $b$ covers the centrifugal and Coriolis effects, $g$ is the generalized gravity vector and $\tau$ represents the joint torque contributions.
$\dot{q}$ and $\ddot{q}$ are the joint velocities and joint accelerations, respectively.
Note that \eqref{eq:dynamics_equation} is a unified equation which applies to both human and object dynamics, hence we drop the superscript $e$ here. Only the expression of the joint torque $\tau$ differs between the human and the object and we give the two expressions next. 

For human, it is the sum of two contributions: the first one corresponds to the internal joint torques (exerted by the muscles for instance) and the second one comes from the contact forces:
\begin{align}
    \tau^\mathrm{h} =
    \begin{pmatrix}
        \mathbf{0}_6 \\ 
        \tau^\mathrm{h}_\mathrm{m}
    \end{pmatrix}
    + \sum_{k=1}^K \left(J^\mathrm{h}_k\right)^Tf_k, \label{eq:human_torque}
\end{align}
where $\tau^\mathrm{h}_\mathrm{m}$ is the human joint torque exerted by muscles, $f_k$ is the contact force at contact point $k$ and $J^\mathrm{h}_k$ is the Jacobian mapping human joint velocities $\dot{q}^\mathrm{h}$ to the Cartesian velocity of contact point $k$ expressed in $k$'s local frame.
Let $n^\mathrm{h}_q$ denote the dimension of $q^\mathrm{h}$, $\dot{q}^\mathrm{h}$ and $\ddot{q}^\mathrm{h}$, then $\tau^\mathrm{h}_\mathrm{m}$ and $J^\mathrm{h}_k$ are of dimension $n_q^h-6$ and $3\times n_q^h$, respectively.
We model the human body and the object as free-floating base systems.
In the case of human body, the six first entries in the configuration vector $q$ correspond to the 6D pose of the free-floating base (translation + orientation), which is not actuated by any internal actuators such as human muscles.
This constraint is taken into consideration by adding the zeros in Eq. \eqref{eq:human_torque}.

In the case of the manipulated object, there is no actuation other than the contact forces exerted by the human.
Therefore, the object torque is expressed as
\begin{align}
    \tau^\mathrm{o} =
    -\sum_{\text{object contact }k} \left(J^\mathrm{o}_k\right)^Tf_k, \label{eq:object_torque}
\end{align}
where the sum is over the object contact points, $f_k$ is the contact force, and $J^\mathrm{o}_k$ denotes the object Jacobian, which maps from the object joint velocities $\dot{q}^\mathrm{o}$ to the Cartesian velocity of the object contact point $k$ expressed in $k$'s local frame.
$J^\mathrm{o}_k$ is a $3\times n^\mathrm{o}_q$ matrix where $n^\mathrm{o}_q$ is the dimension of object configuration vectors $q^\mathrm{o}$, $\dot{q}^\mathrm{o}$ and $\ddot{q}^\mathrm{o}$.

We concatenate the dynamics equations of both human and object to form the overall dynamics in Eq. \eqref{eq:full_body_dynamics} in problem \eqref{eq:general_problem}, and include a \textit{muscle torque} term  
$l^\mathrm{h}_\mathrm{torque} = \|\tau^\mathrm{h}_\mathrm{m}\|^2$ in the overall cost. Minimizing the muscle torque acts as a regularization over the energy consumption of the human body.

\subsection{Enforcing the trajectory smoothness}\label{sec:smoothness}
\paragraph{Regularizing human and object motion.}
Taking advantage of the temporal continuity of video, we minimize the sum of squared 3D joint velocities and accelerations to improve the smoothness of the person and object motion and to remove incorrect 2D poses.
We include the following \textit{motion smoothing} term to the human and object loss in \eqref{eq:general_problem}:
\begin{align}
  l_\mathrm{smooth} = \sum_{j}{\left(\left\|\nu_j(q, \dot{q})\right\|^2 + \left\|\alpha_j(q, \dot{q}, \ddot{q})\right\|^2\right)},
  \label{eq:cost_smooth}
\end{align}
where $\nu_j$ and $\alpha_j$ are the spatial velocity and the spatial acceleration\footnote{Spatial velocities (accelerations) are minimal and unified representations of linear and angular velocities (accelerations) of a rigid body \cite{featherstone2014rigid}. They are of dimension 6.} of joint $j$, respectively.
In the case of object, $j$ represents an endpoint on the object.
By minimizing $l_\mathrm{smooth}$, both the linear and angular movements of each joint/endpoint are smoothed simultaneously.

\paragraph{Regularizing contact motion and forces.}
In addition to regularizing the motion of the joints, we also regularize the contact states and control by minimizing the velocity of the contact points and the temporal variation of the contact force.
This is implemented by including the following \textit{contact smoothing} term in the cost function in problem \eqref{eq:general_problem}:
\begin{align}
    l^\mathrm{c}_\mathrm{smooth} =
    \sum_{j}\sum_{k\in \phi(j)}\delta_j\left(\omega_k\|\dot{c}_k\|^2+\gamma_k\|\dot{f}_k\|^2\right)\mathrm{d}t, \label{eq:contact_smooth}
\end{align}
where $\dot{c}_k$ and $\dot{f}_k$ represent respectively the temporal variation of the position and the contact force at contact point $k$.
$\omega_k$ and $\gamma_k$ are scalar weights of the regularization terms $\dot{c}_k$ and $\dot{f}_k$.
Note that some contact points, for example the four contact points of the human sole during the sole-ground contact, should remain fixed with respect to the object or the ground during the contact phase.
To tackle this, we adjust $\omega_k$ to prevent contact point $k$ form sliding while being in contact. 

\subsection{Optimization}\label{sec:optimization}

\noindent \textbf{Conversion to a numerical optimization problem.}
We convert the continuous problem~\eqref{eq:general_problem} into a discrete nonlinear optimization problem using the collocation approach~\cite{biegler2010nonlinear_chap10}.
All trajectories are discretized and constraints~\eqref{eq:contact_motion_model}, \eqref{eq:full_body_dynamics}, \eqref{eq:force_model} are only enforced on the ``collocation'' nodes of a time grid matching the discrete sequence of video frames.
 The optimization variables are the sequence of human and object poses $[ x_0 ... x_T ]$, torque and force controls $[ u_1 ... u_T ]$, contact locations $[c_0 ... c_T]$, and the scene parameters (ground plane and camera matrix).
The resulting problem is nonlinear, constrained and sparse (due to the sequential structure of trajectory optimization).
We rely on the Ceres solver~\cite{ceres-solver}, which is dedicated to solving sparse estimation problems (e.g.~bundle adjustment~\cite{triggs1999bundle}), and on the Pinocchio software~\cite{carpentier2019pinocchio,pinocchioweb} for the efficient computation of kinematic and dynamic quantities and their derivatives~\cite{carpentier2018analytical}.
Additional details are given in Appendix \ref{appendix:optimization}.

\vspace*{-4mm}
\paragraph{Initialization.} 
Correctly initializing the solver is key to escape from poor local minima. 
We warm-start the optimization by inferring the initial configuration vector $q_k$ at each frame using the human body estimator HMR~\cite{hmrKanazawa18} that estimates the 3D joint angles from a single RGB image. 

\section{Extracting 2D measurements from video}\label{sec:extract_2d_measurements}
In this section, we describe how 2D measurements are extracted from the input video frames during the first, recognition stage of our system.
In particular, we extract the 2D human joint positions, the 2D object endpoint positions and the contact states of human joints.

\vspace*{-4mm}
\paragraph{Estimating 2D positions of human joints.}\label{sec:human_2d_pose}
We use the state-of-the-art Openpose~\cite{cao2017realtime} human 2D pose estimator, which achieved excellent performance on the MPII Multi-Person benchmark~\cite{andriluka20142d}.
Taking a pretrained Openpose model, we do a forward pass on the input video in a frame-by-frame manner to obtain an estimate of the 2D trajectory of human joints, $p^\mathrm{h,2D}_j$.

\vspace*{-4mm}
\paragraph{Recognizing contacts.}\label{sec:contact_recognition}
We wish to recognize and localize contact points between the person and the manipulated object or the ground. 
This is a challenging task due to the large appearance variation of the contact events in the video. 
However, we demonstrate here that a good performance can be achieved  by training a contact recognition CNN module from manually annotated contact data that combine both still images and videos harvested from the Internet. 
In detail, the contact recognizer operates on the 2D human joints predicted by Openpose.
Given 2D joints at video frame $i$, we crop fixed-size image patches around a set of joints of interest, which may be in contact with an object or ground.
Based on the type of human joint, we feed each image patch to the corresponding CNN to predict whether the joint appearing in the patch is in contact or not.
The output of the contact recognizer is a sequence $\delta_{ji}$ encoding the contact states of human joint $j$ at video frame $i$, i.e.\  $\delta_{ji}=1$ if joint $j$ is in contact at frame $i$ and zero otherwise.
Note that $\delta_{ji}$ is the discretized version of the contact state trajectory $\delta_j$  presented in Sec.~\ref{sec:main_stage}.

Our contact recognition CNNs are built by replacing the last layer of an ImageNet pre-trained Resnet model \cite{he2016deep} with a fully connected layer that has a binary output.
We have trained separate models for five types of joints: hands, knees, foot soles, toes, and neck. 
To construct the training data, we collect still images of people manipulating tools using Google image search.
We also collect short video clips of people manipulating tools from Youtube in order to also have non-contact examples.
We run Openpose pose estimator on this data, crop patches around the 2D joints, and annotate the resulting dataset with contact states.

\vspace*{-4mm}
\paragraph{Estimating 2D object pose.}\label{sec:object_2d_endpoints}
The objective is to estimate the 2D position of the manipulated object in each video frame.
To achieve this, we build on instance segmentation obtained by Mask R-CNN~\cite{MaskRCNN}.
We train the network on shapes of object models from different viewpoints and
 apply the trained network on the test videos. The output masks and bounding boxes are used to estimate object endpoints in each frame. The resulting 2D endpoints are used as input to the trajectory optimizer. Details are given next.

In the case of barbell, hammer and scythe, we created a single 3D model for each tool, roughly approximating the shapes of the instances in the videos, and rendered it from multiple viewpoints using a perspective camera.
For spade, we annotated 2D masks of various instances of the tool in thirteen different still images.
The shapes of the rendered 3D models or 2D masks are used to train Mask R-CNN for instance segmentation of each tool.
The training set is augmented by 2D geometric transformations (translation, rotation, scale) to handle the changes in shapes of tool instances in the videos. In addition, domain randomization \cite{loing2018,tobin2017corr} is applied to handle the variance of instances and changes in appearance in the videos caused by illumination: the geometrically transformed shape is filled with pixels from a random image (foreground) and pasted on another random image (background). We utilized random images from the MS COCO dataset \cite{COCO} for this purpose.
We use a Mask R-CNN (implementation \cite{matterport_maskrcnn_2017}) model pre-trained on the MS COCO dataset and re-train the head layers for each tool.

At test time, masks and bounding boxes obtained by the re-trained Mask R-CNN are used to estimate the coordinates of tool endpoints.
Proximity to coordinates of estimated wrist joints
is used to select the mask and bounding box in case multiple candidates are available in the frame.
To estimate the main axis of the object, a line is fitted through the output binary mask.
The endpoints are calculated as the intersection of the fitted line and boundaries of the bounding box.
Using the combination of the output mask and the bounding box compensates for errors in the segmentation mask caused by occlusions.
The relative orientation of the tool (i.e.~the head vs.~the handle of the tool) is determined by spatial location of endpoints in the video frames as well as by their proximity to the estimated wrist joints.


\section{Experiments}\label{sec:results}
In this section we present quantitative and qualitative evaluation of the reconstructed 3D person-object interactions.
Since we recover not only human poses but also object poses and contact forces, evaluating our results is difficult due to the lack of ground truth forces and 3D object poses in standard 3D pose benchmarks such as \cite{h36m_pami}. 
Consequently, we evaluate our motion and force estimation quantitatively on a recent Biomechanics video/MoCap dataset capturing challenging dynamic parkour motions~\cite{maldonado}.
In addition, we report joint errors and show qualitative results on our newly collected dataset of videos depicting handtool manipulation actions.

\subsection{Parkour dataset}\label{sec:parkour_dataset}
This dataset contains videos capturing human subjects performing four typical parkour actions: two-hand jump, moving-up, pull-up and a single-hand hop.
These are highly dynamic motions with rich contact interactions with the environment. The ground truth 3D motion and contact forces are captured with a Vicon motion capture system and several force plates.
The 3D motion and forces are reconstructed with frame rates of $400$Hz and $2200$Hz, respectively, whereas the RGB videos are captured in a relatively lower rate of $25$Hz, making this dataset a challenge for computer vision algorithms due to motion blur.

\vspace*{-4mm}
\paragraph{Evaluation set-up.}
We evaluate both the estimated human 3D motion and the contact forces.
For evaluating the accuracy of the recovered 3D human poses, we follow the common approach of computing the mean per joint position error (MPJPE) of the estimated 3D pose with respect to the ground truth after rigid alignment~\cite{gower1975generalized}.
We evaluate contact forces without any alignment: we express both the estimated and the ground truth 6D forces at the position of the contact aligned with the world coordinate frame provided in the dataset.
We split the 6D forces into linear and moment components and report the average Euclidean distance of the linear force and the moment with respect to the ground truth.

\begin{table}[t]
\small
\centering
\resizebox{\columnwidth}{!}{
\begin{tabular}{lccccc}
\hline
Method                     &   Jump &  Move-up &  Pull-up  &     Hop &    Avg \\ \hline
SMPLify~\cite{bogo2016keep}& 121.75 &   147.41 &   120.48  &  169.36 &  139.69 \\
HMR~\cite{hmrKanazawa18}&    111.36 &   140.16 &   132.44  &  149.64 &  135.65 \\
Ours               & \textbf{98.42} & \textbf{125.21} & \textbf{119.92}& \textbf{138.45}    & \textbf{122.11} \\ \hline
\end{tabular}}
\caption{\small Mean per joint position error (in mm) of the recovered 3D motion for each action on the Parkour dataset.}
\label{tb:mpjpe-galo-per-action}
\vspace*{-1mm}
\end{table}

\begin{table}[t]
\begin{small}
\centering
\resizebox{0.89\columnwidth}{!}{
\begin{tabular}{lcccc}
\hline
                   & L. Sole  & R. Sole         & L. Hand        & R. Hand        \\ \hline
Force (N)     &  144.23        & 138.21         &  107.91        &  113.42   \\
Moment (N$\cdot$m)          & 23.71     & 22.32           &     131.13         & 134.21   \\ \hline
\end{tabular}}
\caption{\small Estimation errors of the contact forces exerted on soles and hands on the Parkour dataset.}
\vspace*{-4mm}
\label{tb:force-errors-galo}
\end{small}
\end{table}

\paragraph{Results.}
We report joint errors for different actions in Table \ref{tb:mpjpe-galo-per-action} and compare the results with the HMR 3D human pose estimator~\cite{hmrKanazawa18}, which is used to warm-start our method.
To make it a fair comparison, we use the same Openpose 2D joints as input.
In addition, we evaluate the recent SMPLify \cite{bogo2016keep} 3D pose estimation method.
Our method outperforms both baselines by more than 10mm on average on this challenging data.
\begin{figure*}[t]
    \centering
    \includegraphics[width=\textwidth]{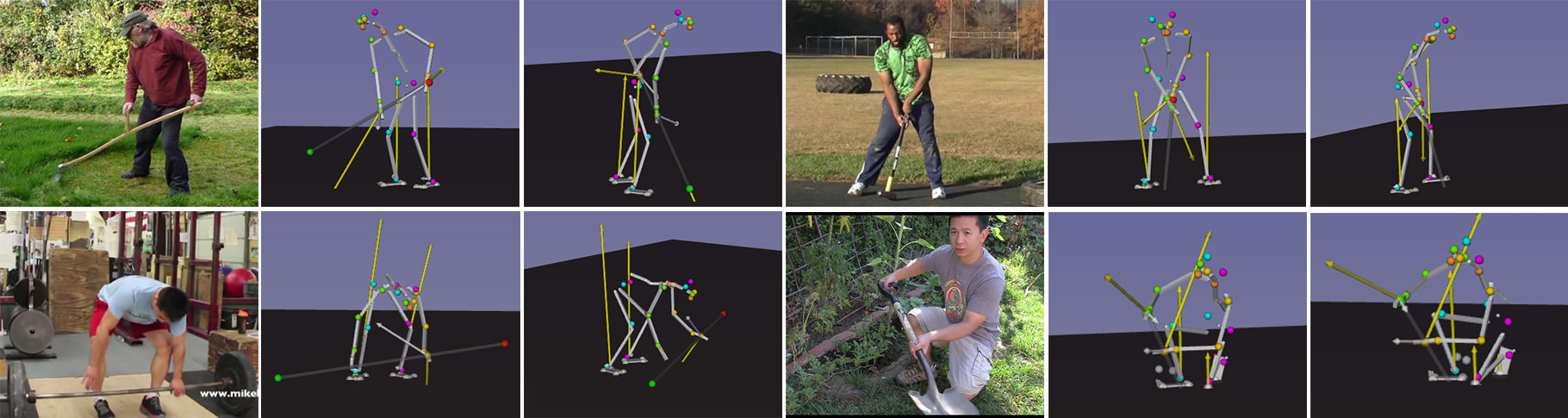}
    \caption{\small Example qualitative results on the Handtool dataset. Each example shows the input frame (left) and two different views of the output 3D pose of the person and the object (middle, right). 
    The yellow and the white arrows in the output show the contact forces and moments, respectively. Note how the proposed approach recovers from these challenging unconstrained videos the 3D configuration of the person-object interaction together with the contact forces and moments. {\bf For additional video results please see the project webpage~\cite{project-page}.}
    }
    \label{fig:qualitative}
\end{figure*}
Finally, Table \ref{tb:force-errors-galo} summarizes the force estimation results. 
To estimate the forces we assume a generic human physical model of mass $74.6$\,kg for all the subjects.
Despite the systematic error due to the generic human mass assumption, the results in Table \ref{tb:force-errors-galo} validate the quality of our force estimation at the soles and the hands during walking and jumping.
We observe higher errors of the estimated moments at hands, which we believe is due to the challenging nature of the Parkour sequences where the entire person's body is often supported by hands.
In this case, the hand may exert significant force and torque to support the body, and a minor shift in the force direction may lead to significant errors.

\subsection{Handtool dataset}\label{sec:handtooldataset}
In addition to the Parkour data captured in a controlled set-up, we would like to demonstrate generalization of our approach to the ``in the wild'' Internet instructional videos.
For this purpose, we have collected a new dataset of object manipulation videos, which we refer to as the Handtool dataset.
The dataset contains videos of people manipulating four types of tools: {\em barbell}, {\em hammer}, {\em scythe}, and {\em spade}.
For each type of tool, we chose among the top videos returned by YouTube five videos covering a range of actions.
We then cropped short clips from each video showing the whole human body and the tool.

\vspace*{-2mm}
\paragraph{Evaluation of 3D human poses.}
For each video in the Handtool dataset, we have annotated the 3D positions of the person's left and right shoulders, elbows, wrist, hips, knees, and ankles, for the first, the middle, and the last frame.
We evaluate the accuracy of the recovered 3D human poses by computing their MPJPE after rigid alignment.
Quantitative evaluation of the recovered 3D poses is shown table \ref{tb:mpjpe-handtools}.
On average, our method outperforms the strong HMR~\cite{hmrKanazawa18} and SMPLify~\cite{bogo2016keep} baselines. However, the differences between the methods are reaching the limits of the accuracy of the manually provided 3D human pose annotations on this dataset. Videos available at project website~\cite{project-page} demonstrate that our model produces smooth 3D motion, respecting person-object contacts and capturing the interaction of the person with the tool. This is not the case for the HMR~\cite{hmrKanazawa18} and SMPLify~\cite{bogo2016keep} baselines that are applied to individual frames and do not model the interaction between the person and the tool. Example results for our method are shown in Figure~\ref{fig:qualitative}.   
For additional results including examples of the main failure modes please see Appendix \ref{appendix:qualitive_results}.

\vspace*{-2mm}
\paragraph{Evaluation of 2D object poses.}
To evaluate the quality of estimated object poses, we manually annotated 2D object endpoints in every 5th frame of each video in the Handtool dataset and calculated the 2D Euclidean distance (in pixels) between each manually annotated endpoint and its estimated 2D location provided by our method.
The 2D location is obtained by projecting the estimated 3D tool position back to the image plane. 
We compare our results to the output of the Mask R-CNN instance segmentation baseline \cite{MaskRCNN} (which provides initialization for our person-object interaction model).
In Table~\ref{table:evaluation2dposeall2550100} we report for both methods the percentage of endpoints for which the estimated endpoint location lies within 25, 50, and 100 pixels from the annotated ground truth endpoint location. 
The results demonstrate that our approach provides more accurate and stable object endpoint locations compared to the Mask R-CNN baseline thanks to modeling the interaction between the object and the person.

\begin{table}[t]
\footnotesize
\centering
\resizebox{\columnwidth}{!}{
\begin{tabular}{lcccccc}
\hline
Method                       & Barbell        & Spade         & Hammer         & Scythe         & Avg            \\ \hline
SMPLify~\cite{bogo2016keep}  & 130.69         & 135.03        & \textbf{93.43}          & \textbf{112.93}         & 118.02         \\
HMR~\cite{hmrKanazawa18}     & 105.04         & 97.18         & 96.34          & 115.42         & 103.49         \\
Ours               & \textbf{104.23} & \textbf{95.21}&        95.87 & 114.22 & \textbf{102.38} \\ \hline
\end{tabular}}
\caption{\small Mean per joint position error (in mm) of the recovered 3D human poses for each tool type on the Handtool dataset.}
\label{tb:mpjpe-handtools}
\end{table}

\begin{table}[t]
\small
\centering
\resizebox{\columnwidth}{!}{
\begin{tabular}{lcccc}
\hline
Method                     & Barbell         & Spade        & Hammer          & Scythe       \\ \hline
Mask R-CNN \cite{MaskRCNN} & 33/42/54       & 54/79/93     & 35/44/45    & 63/72/76    \\
Ours                       & \textbf{38/71/98}   & \textbf{57/86/99}   & \textbf{61/91/99}    & \textbf{69/88/98} \\ \hline
\end{tabular}}
\caption{\small The percentage of endpoints for which the estimated 2D location lies within 25/50/100 pixels (in 600$\times$400 pixel image) from the manually annotated ground truth location. }
\vspace*{-4mm}
\label{table:evaluation2dposeall2550100}
\end{table}

\vspace*{-1mm}
\section{Conclusion}
\vspace*{-2mm}
We have developed a visual recognition system that takes as input video frames together with a simple object model, and outputs a 3D motion of the person and the object including contact forces and torques actuated by the human limbs. 

We have validated our approach on a recent MoCap dataset with ground truth contact forces.
Finally, we have collected a new dataset of unconstrained instructional videos depicting people manipulating different objects and have demonstrated benefits of our approach on this data. 
Our work opens up the possibility of large-scale learning of human-object interactions from Internet instructional videos~\cite{Alayrac16unsupervised}.

\paragraph{Acknowledgments.}
We warmly thank Bruno Watier (Universit\'e Paul Sabatier and LAAS-CNRS) and Galo Maldonado (ENSAM ParisTech) for setting up the Parkour dataset.
This work was partly supported by the ERC grant
LEAP (No.\,336845), the H2020 Memmo project, CIFAR Learning in Machines\&Brains program, and the European Regional Development Fund under the project IMPACT (reg. no. CZ.02.1.01/0.0/0.0/15\_003/0000468).

{\small
\bibliographystyle{ieee}
\bibliography{egbib}
}

\clearpage
\appendix

\section*{Outline of the appendices}
In these appendices, we provide additional technical details and qualitative results of the proposed method.
In appendix~\ref{appendix:human_object_models}, we provide a comprehensive description of the parametric human and object model we use for the trajectory optimization.
Then, in appendix~\ref{appendix:ground_force_generators} we give details of the ground contact force generators mentioned in the main paper (section~\ref{sec:physical_plausibility}).
In appendix~\ref{appendix:optimization}, we present additional optimization details concerning the trajectory estimation stage.
Finally, in section~\ref{appendix:qualitive_results}, we present additional qualitative results including several typical failure modes of our method.

\section{Parametric human and object models}
\label{appendix:human_object_models}
\begin{figure}[t]
    \centering
    \includegraphics[width=0.45\textwidth]{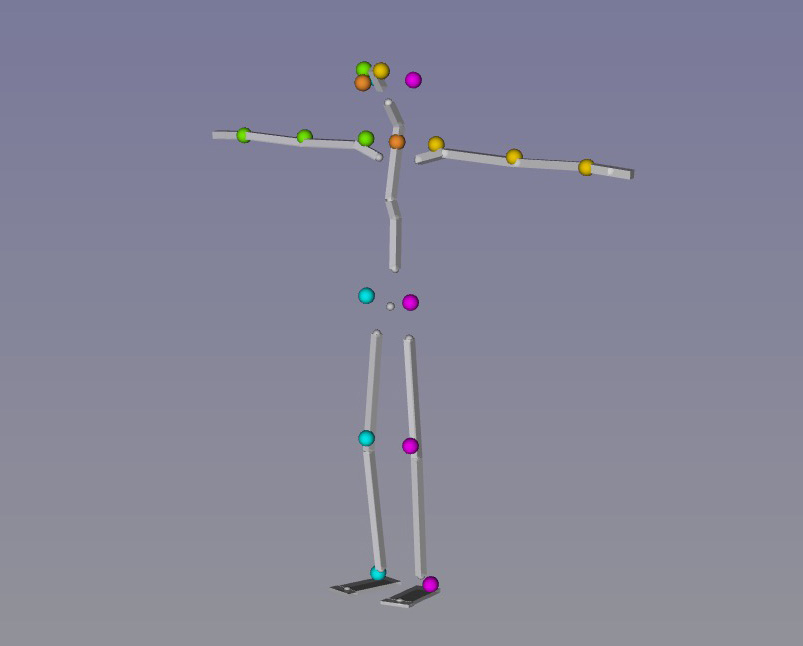}
    \caption{\small Our human model in reference posture. The skeleton consists of 1 free-floating basis joint corresponding to pelvis, and 23 spherical joints. The colored spheres are 18 virtual markers that correspond to 18 Openpose joints. Each marker is associated to a semantically corresponding joint in our model.}
    \label{fig:human_model}
\end{figure}

\paragraph{Human model.} 
We model the human body as a multi-body system consisting of a set of rotating joints and rigid links connecting them.
We adopt the joint definition of the SMPL model \cite{loper2015smpl} and approximate the human skeleton as a kinematic tree with 24 joints: one free-floating joint and 23 spherical joints.
Figure \ref{fig:human_model} illustrates our human model in a canonical pose.
A free-floating joint consists of a 3-dof translation in $\mathbb{R}^3$ and a 3-dof rotation in $SO(3)$; we model the pelvis by a free-floating joint to describe the person's body orientation and translation in the world coordinate frame.
A spherical joint is a 3-dof rotation; it represents the relative rotation between two connected links in our model.
In practice, we use unit quaternions to represent 3D rotations and axis-angles to describe angular velocities.
As a result, the configuration vector of our human model $q^\mathrm{h}$ is a concatenation of the configuration vectors of the 23 spherical joints (dimension 4) and the free-floating pelvis joint (dimension 7), hence of dimension 99.
The corresponding human joint velocity $\dot{q}^\mathrm{h}$ is of dimension $23\times 3+6=75$ (by replacing the quaternions with axis-angles).
For simplicity, in the main paper we do not distinguish this dimension difference and consider both $q^\mathrm{h}$ and $\dot{q}^\mathrm{h}$ to be represented using axis-angles, hence of the same dimension $n_q^\mathrm{h}=75$.
In addition, based on these 24 joints, we define 18 ``virtual markers'' (shown as colored spheres in Figure \ref{fig:human_model}) that represent the 18 Openpose joints.
These markers are used instead of the 24 joints to compute the re-projection errors with respect to the Openpose 2D detections.

\begin{figure}[t]
    \centering
    \includegraphics[width=0.45\textwidth]{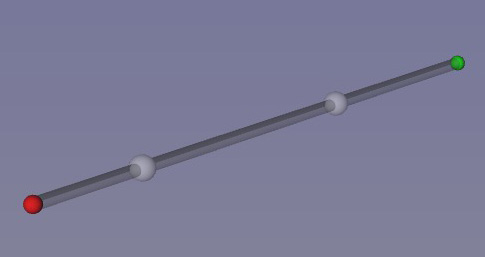}
    \caption{\small All four handtools are represented by a single object model shown in this image.
    The object model consists of 1 free-floating basis joint corresponding to the handle end (red sphere), 1 prismatic joint corresponding to the head of the tool (green sphere), and several prismatic joints corresponding to the location of the contact points (grey translucent spheres in the middle).
    The contact points should lie on the feasible contact area (grey stick) formed by the two endpoints.}
    \label{fig:object_model}
\end{figure}

\paragraph{Object models.}
All four objects, namely barbell, hammer, scythe and spade, are modeled as non-deformable rigid sticks.
The configuration $q^\mathrm{o}$ represents the 6-dof displacement of the stick handle, as illustrated in Figure \ref{fig:object_model}.
In practice, $q^\mathrm{o}$ is a 7-dimensional vector containing the 3D translation and 4D quaternion rotation of the free-floating handle end.
The object joint velocity $\dot{q}^\mathrm{o}$ is of dimension 6 (by replacing the quaternion with an axis-angle).
The handtools that we are modelling have the stick handle as the contact area. 
We ignore the handle's thickness and represent the contact area using the line segment between the two endpoints of the handle.
Depending on the number of human joints in contact with the object, we associate the same number of contact points to the object's local coordinate frame.
These contact points can be located at any point along the feasible contact area.
In practice, all object contact points together with the endpoint corresponding to the head of the handtool are implemented as ``virtual'' prismatic joints of dimension 1.


\section{Generators of the ground contact forces}
\label{appendix:ground_force_generators}

In this section, we describe the generators $g^{(3)}_n$ and $g^{(6)}_{kn}$ for computing the contact forces exerted by the ground on the person.
Recall from the main paper that we consider different contact models depending on the type of the joint. We model the planar contacts between the human sole and the ground by fitting the point contact model (given by eq.~\eqref{eq:point_contact} in the main paper) at each of the four sole vertices.
For other types of ground contacts, e.g.~the knee-ground contact, we apply the point contact model directly at the human joint.
We model the ground as a 2D plane $G = \{p\in \mathbb{R}^3|a^Tp=b\}$ with a normal vector $a\in \mathbb{R}^3$, $a\neq 0$, $b\in \mathbb{R}$ and a friction coefficient $\mu$.
In the following, we first provide the expression of the 3D generators $g^{(3)}_n$ for modeling point contact forces and then derive the 6D generators $g^{(6)}_{kn}$ for modeling planar contact forces.

\paragraph{3D generators $g^{(3)}_n$ for point contact forces.}
Let $p_k$ be the position of a contact point $k$ located on the ground surface, i.e. $a^Tp_k=b$.
We define at contact point $k$ a right-hand coordinate frame $C$ whose $xz$-plane overlaps the plane $G$ and whose $y$-axis points towards the gravity direction, i.e., the opposite direction to the ground normal $a$.
During point contact, it is a common assumption that the ground exerts only linear reaction forces on the contact point $c$.
In other words, the spatial contact force expressed in the local frame $C$ can be expressed as
\begin{align}
    ^C\phi = 
    \begin{pmatrix}
        f \\ 
        \mathbf{0}_{3\times1}
    \end{pmatrix}, \label{eq:point_contact_force}
\end{align}
where the linear component $f$ must lie in the second-order cone $\mathcal{K}^3 = \{f=(f_x,f_y,f_z)^T|\sqrt{f_x^2 + f_z^2} \leq -f_y \tan\mu\}$, which can be approximated by the pyramid ${\mathcal{K}^3}^\prime = \{f=\sum_{n=1}^4{\lambda_n g^{(3)}_n}|\lambda_n\geq 0\}$, with a set of 3D-generators
\begin{align}
    g^{(3)}_1 &= 
    \left(\sin{\mu}, -\cos{\mu}, 0\right)^T, \\
    g^{(3)}_2 &= 
    \left(-\sin{\mu}, -\cos{\mu}, 0\right)^T, \\
    g^{(3)}_3 &= 
    \left(0, -\cos{\mu}, \sin{\mu}\right)^T, \\
    g^{(3)}_4 &= 
    \left(0, -\cos{\mu}, -\sin{\mu}\right)^T.
\end{align}
More formally, we are approximating the friction cone $\mathcal{K}^3$ with the conic hull ${\mathcal{K}^3}^\prime$ spanned by 4 points on the boundary of $\mathcal{K}^3$, namely, $g^{(3)}_n$ with $n=1,2,3,4$. 

\paragraph{6D generators $g^{(6)}_{kn}$ for planar (sole) contact forces.}
Here we show how to obtain the 6D generator $g^{(6)}_{kn}$ from $g^{(3)}_{n}$ and the contact point position $p_k$.
As described in the main paper, we approximate human sole as a rectangle area with 4 contact points.
We assume that the sole overlaps the ground plane $G$ during contact.
Similar to the point contact, we define 5 parallel coordinate frames $C_k$, one at each of the four sole contact points, plus a frame $A$ at the ankle joint.
Note that the frames $C_k$ and $A$ are parallel to each other, i.e., there is no rotation but only translation when passing from one frame to another.
We can write the contact force at contact point $k$ as the 6D spatial force
\begin{align}
    ^{C_k}\phi_k =
    \sum_{n=1}^4 \lambda_{kn}
    \begin{pmatrix}
        g^{(3)}_n \\ 
        \mathbf{0}_{3\times1}
    \end{pmatrix}
    , \text{ with } \lambda_{kn} \geq 0.
\end{align}
We denote by $^Ap_k$ the position of contact point $c_k$ in the ankle frame $A$, and by $^AX_{C_k}^*$ the matrix converting spatial forces from frame $C_k$ to frame $A$.
We can then express the contact force in frame $A$:
\begin{align}
    ^A\phi &= \sum_{k=1}^4{{^AX_{C_k}^*}^{C_k}\phi_k} \\
    &= \sum_{k=1}^4{\begin{pmatrix}
        I_3 & ^Ap_k\times\\
        0_3 & I_3\\
    \end{pmatrix}^{-T}{^{C_k}\phi_k}} \\
    &=\sum_{k=1}^4\sum_{n=1}^4 \lambda_{kn} g^{(6)}_{kn}, \label{eq:planar_contact_force_6d}
\end{align}
where 
\begin{align}
    g^{(6)}_{kn}=
    \begin{pmatrix}
        g^{(3)}_n \\ 
        ^Ap_k\times g^{(3)}_n
    \end{pmatrix}.
\end{align}

\section{Optimization details}
\label{appendix:optimization}

\paragraph{Discretizing the original problem.}
As described in the main paper, we discretize the motion and control trajectories $\underline{x}$, $\underline{u}$ and $\underline{c}$, and enforce the constraints \eqref{eq:contact_motion_model}, \eqref{eq:full_body_dynamics}, \eqref{eq:force_model} on the time samples corresponding to the input video frames.
We replace the integral in the objective function by a sum over video frames, and rewrite the cost and constraint terms which include derivatives of the state (e.g. joint accelerations) by approximating the derivatives with the backward finite difference scheme (e.g. $a_t := (v_{t} - v_{t-1}) / \Delta t $, with $\Delta t$ the duration between the two video frames).

\paragraph{Problem sparsity.}
The problem after discretization becomes a large, sparse and non-linear optimization problem.
This is because the discretized objective function becomes a sum of terms that each depend on one time sample (denoted by $i$) and a subset of the variables $[x_i,u_i,c_i]$ corresponding to $i$.
Only a few regularization terms, e.g. the motion smoothing term \eqref{eq:cost_smooth} and the contact smoothing term \eqref{eq:contact_smooth}, may depend on two or three successive frames.
The problem sparsity is important to take into account, as it significantly reduces the complexity of computation from $\mathcal{O}(T^3)$ (without sparsity) to $\mathcal{O}(T)$ (using the problem sparsity). 

\paragraph{Solving the problem.}
We solve the problem using the Levenberg-Marquardt algorithm which is known to be effective in solving non-linear least squares problems.
We use the Ceres solver~\cite{ceres-solver}. 
As the solver only allows to define bound constraints, we further implement our nonlinear constraints as penalties in the cost function.
The bound constraints in problem~\eqref{eq:general_problem} in the main paper, such as the non-negative coefficients of the force generators $\lambda_{jkn} \geq 0$, are kept as hard constraints in our implementation.
The optimization takes 3.2 seconds per frame on average on a common desktop machine.

\paragraph{Multi-stage optimization.} Solving the optimization problem all at once would lead to poor local minima. Instead, we solve a cascade of sub-problems composed of three stages.
In the first stage, we solve the problem~\eqref{eq:general_problem} in the main paper only for the kinematics variables ($q,\dot q, \ddot q$) by ignoring the dynamics constraints given by equations~\eqref{eq:full_body_dynamics} and~\eqref{eq:force_model} in the main paper. 
In this stage, the torque vectors and the contact forces are not decision variables of the optimization problem.
Note that the location of the manipulated object varies significantly across the Handtool dataset.
To address this, we try four initialization options with different pre-defined 3D object orientations.
We run this stage for each initialization and  pick among the four resulting models the one with the lowest cost.
In the second stage, we fix the values of the configuration vectors and their time derivatives, and optimize only for the torque control $\tau$ input and the contact forces $f_k$.
In the last stage, we solve for the complete set of kinematics and control variables all at once, starting from the values provided by the two first stages.
It would be then possible to continue improving the solution by pursing the aforementioned  alternative descent scheme, but we found that a single pass was already sufficient to obtain good qualitative results.

\section{Additional qualitative results}
\label{appendix:qualitive_results}
\begin{figure*}[t]
    \centering
    \includegraphics[width=\textwidth]{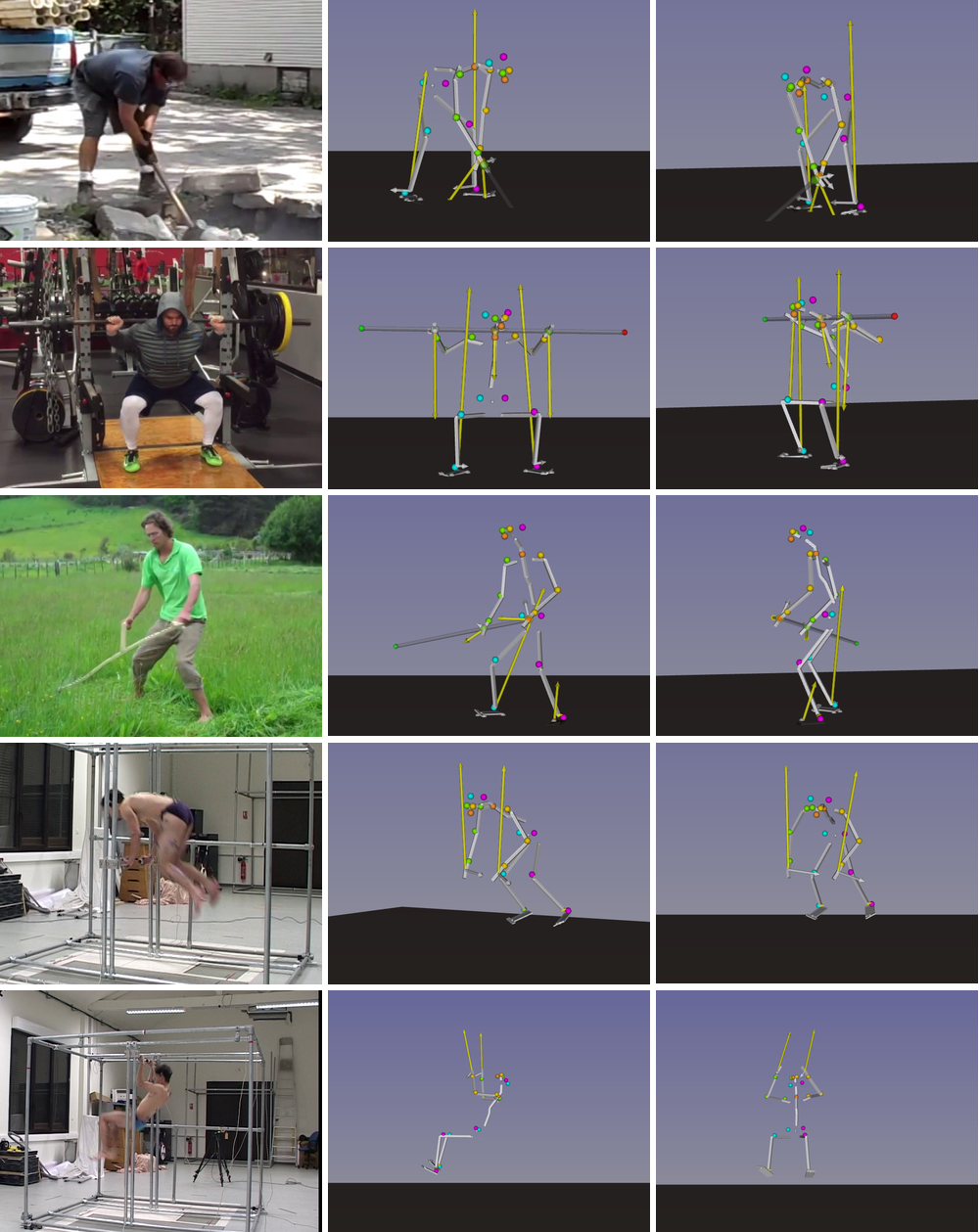}
    \caption{\small {\bf Additional qualitative results} for the Handtool (top three rows) and the Parkour (bottom two rows) datasets at selected frames.
    Each row shows the input video frame (left), the output 3D pose of the person and the object from the original viewpoint (middle) and the same output from a different viewpoint (right). 
    Yellow and white arrows in the output show the contact forces and moments, respectively. Note that in the Parkour dataset we recognize the contact states of individual limbs but do not recognize and model the pose of the object (the metal construction) the person is interacting with. 
    }
    \label{fig:qualitative-appendix}
\end{figure*}

Here we provide additional qualitative results of our method on the Handtool and Parkour datasets (described respectively in Sections \ref{sec:handtooldataset} and  \ref{sec:parkour_dataset}).
Each row of Figure~\ref{fig:qualitative-appendix} shows results at a selected frame from one of the two datasets.
For each sample, we first show the original input frame (left column), followed by the estimated 3D person-object interaction shown from the original viewpoint (middle column), and the same 3D scene from a different viewpoint (right column).
Note how our method produces realistic 3D human and object poses as well as reasonable contact forces from a single ``in-the-wild'' instructional video from the Internet.
The linear forces and moments are visualized by 3D arrows in yellow and white color, respectively.
The length of the arrow represents the magnitude of the force normalized by gravity.

Regarding 3D pose estimation, the results show that our model resolves the depth ambiguity by enforcing the contact motion and force models introduced in Section \ref{sec:physical_plausibility}.
For example, in the hammering video in the first row, the person's hands holding the hammer are restricted to be on the handle, thus reducing the depth ambiguity.
For the barbell and scythe actions where the person-object contacts are considered fixed, our contact model preserves the distance between the person's hands when the person turns around.

The qualitative results also demonstrate that our model predicts reasonable contact forces.
The directions of the contact forces exerted on the person's hands are consistent with the object's motion trajectory and gravity, and the ground reaction forces generally point towards the direction opposite to gravity.
Specifically, in the video with the person practicing back squat with barbell (see the second row of Figure~\ref{fig:qualitative-appendix}), the reconstructed object contact forces and ground reaction forces are distributed evenly on the person's hands, and feet, respectively. {\bf Please see video examples on the project website~\cite{project-page}.} 

\begin{figure}[ht]
    \centering
    \includegraphics[width=0.49\textwidth]{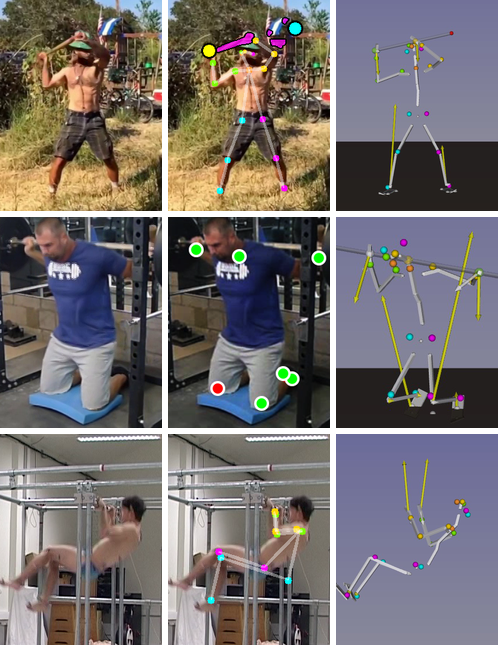}
    \caption{\small {\bf Main failure modes of our method.} The main failure modes are: (i) Incorrectly detected object (top row): the handle of the sledge-hammer  (shown  as  cyan  dot  in  the  image)  is  incorrectly detected, which affects the 3D output of our model. (ii) Contact recognition errors (middle row): the person's right knee is incorrectly recognized as not in contact (red), leading to incorrect force estimation.
    (iii) Incorrectly localized human joints (bottom row): the missing 2D detection of the person's left foot has lead to  errors in estimating the 3D location of the left leg. }
    \label{fig:failure_modes}
\end{figure}

\paragraph{Failure modes.}
Figure~\ref{fig:failure_modes} illustrates the main failure modes of the proposed method, which include: (i) undetected or incorrectly detected objects (top row), (ii) contact recognition errors (middle row) and (iii) incorrectly localized human joints in the image (bottom row).

\end{document}